%% file: root_aa.tex
\documentclass[letterpaper, 10 pt, conference]{ieeeconf}  

\IEEEoverridecommandlockouts                              

\title{\LARGE \bf
Aggressive Online Control of a Quadrotor via Deep Network Representations of Optimality Principles
}

\usepackage{amsmath}
\usepackage[]{algorithm}
\usepackage{algpseudocode}
\usepackage{graphicx}
\usepackage{subfigure}
\usepackage{mathtools}
\usepackage{color}

\author{Shuo Li$^{1}$, Ekin \"Ozt\"urk$^{2}$, Christophe De Wagter$^{1}$, Guido C. H. E. de Croon$^1$, Dario Izzo$^{2}$ 
\thanks{*This work was supported by ESA}
\thanks{$^{1}$ The authors are with the Micro Air Vehicle Lab of the Faculty of Aerospace Engineering, Delft University of Technology, 2629 HS Delft, The Netherlands (e-mail: s.li-4@tudelft.nl; c.deWagter@tudelft.nl; g.c.h.e.decroon@tudelft.nl)}%
\thanks{$^{2}$ The authors are with the Advanced Concepts Team, European Space Agency, Keplerlaan 1, 2201 AZ, Noordwijk, NL. (e-mail: ekin.ozturk@esa.int;dario.izzo@esa.int}%
}

\newcommand{\norm}[1]{\left\lVert#1\right\rVert}

\begin{document}

\maketitle
\thispagestyle{empty}
\pagestyle{empty}

\begin{abstract}


Optimal control holds great potential to improve a variety of robotic applications.
The application of optimal control on-board limited platforms has been severely hindered by the large computational requirements of current state of the art implementations. In this work, we make use of a deep neural network to directly map the robot states to control actions. The network is trained offline to imitate the optimal control computed by a time consuming direct nonlinear method. A mixture of time optimality and power optimality is considered with a continuation parameter used to select the predominance of each objective. 
We apply our networks (termed G\&CNets) to aggressive quadrotor control, first in simulation and then in the real world. We give insight into the factors that influence the `reality gap' between the quadrotor model used by the offline optimal control method and the real quadrotor. Furthermore, we explain how we set up the model and the control structure on-board of the real quadrotor to successfully close this gap and perform time-optimal maneuvers in the real world. Finally, G\&CNet's performance is compared to state-of-the-art differential-flatness-based optimal control methods. We show, in the experiments, that G\&CNets lead to significantly faster trajectory execution due to, in part, the less restrictive nature of the allowed state-to-input mappings. 

\end{abstract}


\section{INTRODUCTION}


\PARstart{A}{} major challenge in the field of drone control is to achieve aggressive (high-speed) autonomous flight. In terms of control, much research focuses on designing controllers which can track a reference guidance trajectory also when considering unmodeled dynamics, nonlinearities and disturbances which become significant when the maneuver of the drone gets aggressive \cite{smeur2015adaptive,tal2018accurate}. In terms of guidance, multiple methods varying from a simple setpoint to high order polynomial trajectory generation methods have shown their feasibility in guiding a quadrotor to the desired target including some time optimality principles.

Two fundamentally different approaches are used to obtain aggressive quadrotor trajectories. The first one is differential flatness based trajectory generation and control \cite{mellinger2011minimum,bry2015aggressive}.
This method is able to generate aggressive trajectories for quadrotors (based on a minimum-time polynomial guidance), and hence it is widely used in real quadrotor flights. However, 
the resulting trajectory can be far from being truly time optimal.

The second approach uses optimal control theory to find and fly a trajectory that incorporates the required optimality principles. Due to the time-consuming nature of this calculation, this method is unsuitable for an online implementation \cite{hehn2012performance,morbidi2016minimum}.
Several methods have been proposed to address this,
where the most common is to represent the system dynamics as a series of simpler linear systems with analytical solutions
\cite{hehn2011quadrocopter,santos2016optimized}. Unfortunately, this simplification can lead to an inaccurate representation of the nonlinear response of the system and can thus negatively impact performance.
An alternative approach is to find and use, on-board, a sub-optimal solution instead.
For example, by using the result of the first iteration of a nonlinear programming (NLP) solver \cite{geisert2016trajectory} which, although incomplete, is faster to compute.

In recent years, leveraging significant advances in machine learning techniques and in particular in artificial neural networks, a number of new methods have been proposed relevant to the aggressive control of quadrotor trajectories. Reference trajectories have been optimized using DNNs \cite{li2017deep}, waypoint tracking has been achieved by means of reinforcement learning \cite{hwangbo2017control} and trajectory tracking using RBFNN \cite{li2016adaptive}. Tang et al. \cite{tang2018learning} combine both optimal control and machine learning. Their experimental results have shown that
a trained neural network can predict an optimal trajectory to the target point, which can then be tracked using PID control.
This work is an important step towards online optimal control, however the main computation is done on a workstation (i.e. not on-board) and, since a PID controller is introduced to track the reference, there are delays during the tracking as a result of which the controls may violate the constraints due to the feedback term. In a different context (i.e. spacecraft landing and mass optimal control) Sanchez et al. \cite{sanchez2018real} successfully introduced the use of imitation learning of optimal controls to train DNNs capable of safely steering the system to desired target positions. Following that work, Tailor and Izzo \cite{Tailor2019} made an extensive study of the technique on simulated drone dynamics and Izzo et al. \cite{izzo2018stability} introduced the term G\&CNets (guidance and control networks) to refer to these networks and showed how to study the stability of the resulting trajectories analytically via differential algebraic techniques.

In this letter, extending previous work on G\&CNets, we present an approach for the on-board optimal control problem of a quadrotor that does not need a PID controller to track the trajectory and we test it during real flights. In our approach, 250,000 optimal trajectories are generated offline. Then, a G\&CNet---which is a neural network trained to learn this dataset---is computed. Instead of predicting an optimal trajectory as the work in \cite{tang2018learning}, G\&CNet predicts the required optimal thrust directly which will be transferred to the optimal pitch rate acceleration and sent to the controller, and thus can be seen in the context of non-Linear MPC. Since the work of \cite{tang2018learning} is difficult to reproduce, we made the comparison between G\&CNet and the differential flatness based trajectory generation and control (DiffG\&C) in simulation. The simulation results show that the proposed G\&CNet can guide the drone to the target points much faster while satisfying optimality principles. Finally, the developed G\&CNet and DiffG\&C controllers are verified in real-world flight tests where the results show that the on-board G\&CNet can guide the drone to the target with a resulting real-time trajectory that is very close to the theoretical optimal solution.

\section{DESIGN OF THE G\&CNET}\label{sec:optimalcontrol}

\subsection{The dynamical system}

\begin{figure}[hbt]
    \centering
  \def\svgwidth{0.5\columnwidth}
  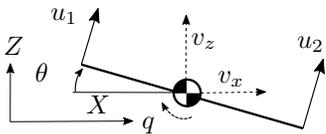

    \caption{Axis definition}
    \label{fig:axisdef}
\end{figure}

Specifying the state of a quadrotor in the $xoz$ plane as

\begin{align}
    \mathbf{x} = 
    \left[x\quad z\quad v_x\quad v_z\quad\theta\quad q\right]
\end{align}

as defined in Fig.~\ref{fig:axisdef}, the dynamical model for which we compute the optimal control is:

\begin{align}
\mathbf{f}(\mathbf{x},\mathbf{u}) =
\left[
\begin{array}{l}
\dot x = v_x \\
\dot z = v_z \\
\dot v_x = - \left[u_\Sigma \frac{{\Delta F}}{m} + 2\frac{\underline F}m \right] \sin \theta  -\beta v_x \\
\dot v_z = \left[u_\Sigma \frac{{\Delta F}}{m} + 2\frac{\underline F}m \right] \cos \theta -g_0 -\beta v_z \\
\dot \theta = q \\
\dot q = \frac{L}{I_{xx}}{\Delta F}(u_2 - u_1)
\end{array}
\right]
\label{equ:simulation model}
\end{align}

where $\Delta F=\overline F - \underline F=0.59\ \mathrm N$ is the range of the thrust magnitude, $\overline F = 2.35\ \mathrm N$ is the maximum thrust, $\underline F = 1.76\ \mathrm N$ is the minimum thrust, $\beta = 0.5$ is the drag coefficient, $m = 0.389\ \mathrm{kg}$ is the quadrotor mass, $L = 0.08\ \mathrm m$ is the length of the quadrotor, $I_{\text{xx}} = 0.001242\ \mathrm{kg}\ \mathrm{m^{2}}$ is the moment of inertia about the x-axis, $\mathbf{u}=\left[u_1, u_2\right]\in\left[0,1\right]$ are the left and right throttles respectively, and $u_\Sigma=(u_1 + u_2)$.

\subsection{The optimisation problem}

The cost function we need to minimise for the optimal controls is:

\begin{align}
    J(\epsilon,t_f,\mathbf{u}(t))=(1-\epsilon) t_f + \epsilon\int_{0}^{t_f}{(u_1(t)^2 + u_2(t)^2)\mathrm{d}t}
\label{equ:cost function}
\end{align}

where $\epsilon\in\left[0,1\right]$ is a hybridisation parameter. When $\epsilon=0$, the cost function is exactly time-optimal, and when $\epsilon=1$, the cost function is exactly power-optimal. With this parameter we are able to generate datasets from time-optimal to power-optimal continuously. Similar to the weighting factor of \cite{tailor2019imitation}, we set $\epsilon$ close to zero ($\epsilon=0.2$) to improve the numerical convergence of the problem and avoid difficult to track control profiles. We trained two networks for $\epsilon=0.5$ and $\epsilon=0.2$ in order to compare how well the quadrotor is able to track and execute the optimal controls with differing degrees of aggressiveness. As we are more interested in time-optimal guidance and control, the dataset and training details focus only on the $\epsilon=0.2$ controller, but the same arguments and methods apply to the $\epsilon=0.5$ controller.

\begin{equation}
\begin{aligned}
\underset{\mathbf{u},t_f}{\text{minimize}} \quad & J(\epsilon,t_f,\mathbf{u}(t))\\
\text{subject to} \quad & \mathbf{\dot x} = \mathbf{f}(\mathbf{x},\mathbf{u}), \ \forall t \\
 &\mathbf{x}(0) = \mathbf{x}_0  \\   
 &   \mathbf{x}(t_f) = \mathbf{0} \\ 
\end{aligned}
\label{equ:optimisation problem}
\end{equation}


Using a direct transcription and collocation method (Hermite-Simpson transcription), the trajectory optimisation problem is transformed into an NLP problem \cite{tailor2019imitation}. The AMPL modelling language was used to specify the NLP problem which was then solved via SNOPT, an SQP NLP solver. Solving for 250,000 trajectories with initial states, $\mathbf{x}_0$, drawn uniformly from $x_0\in[-10,10]\ \mathrm{m}$, $z_0\in[-10,10]\ \mathrm{m}$, $v_{x0}\in[-5,5]\ \mathrm{m\ s^{-1}}$, $v_{z0}\in[-5,5]\ \mathrm{m\ s^{-1}}$, $\theta_0\in[-\pi/3,\pi/3]\ \mathrm{rad}$, and $\omega_0\in[-0.01,0.01]\ \mathrm{rad\ s^{-1}}$, we obtain a database of state-control pairs of the form:
\begin{equation}
\begin{aligned}
    & \kappa_i = \left(\mathbf{x}^{(i)}_j,\mathbf{u}^{(i)}_j\right)_{j=1}^K\quad {\mathrm{where}} & \\ 
    & \mathbf{x}^{(i)}_1=\mathbf{x}^{(i)}_0,\mathbf{x}^{(i)}_J=\mathbf{0}\quad i=1,...,M & \\
\end{aligned}
\end{equation}
where $i$ indexes the trajectories and $K=81$ is the number of grid points in the Hermite-Simpson transcription \cite{tailor2019imitation}. We solved for 250,000 trajectories of which 214,210 converged, and following an 80-10-10 split, these trajectories were split into training, validation and test sets. Overall, this translates to 13,880,808 state-control pairs that the network was trained on, and 1,735,101 that the network was tested on.

\subsection{Network architecture and training}

We construct neural network architectures in the same manner as \cite{tailor2019imitation} with 3 layers, 100 hidden units with softplus activation functions, and sigmoid activation functions for the output controls.

Thus we train on the loss function:
\begin{equation}
    l = \norm{\mathcal{N}(\mathbf{x}) - \mathbf{u^*}}^2
\end{equation}
with a minibatch size of 256 and a starting learning rate of $10^{-3}$ using the Adam optimizer. For further details on network training and construction, refer to \cite{tailor2019imitation}. From this training, the $\epsilon=0.2$ network was able to achieve a mean absolute error (MAE) of $0.0105$ for $u_1$ and $0.0107$ for $u_2$ on the training set, and a MAE of $0.0108$ for $u_1$ and $0.0109$ for $u_2$ on the test set.

\section{SIMULATION RESULT AND ANALYSIS}
\label{sec:SIMULATION RESULT AND ANALYSIS}
In this section, we analyse the theoretical performance of the proposed optimal controller. First we discuss the simulated stability characteristics of the G\&CNet($\epsilon=0.2$) controller. Then we introduce the aforementioned DiffG\&C as a benchmark controller. Finally, we detail the simulation of both methods and present a comparison between simulations.

\subsection{Stability of Neural Network Controller}

\begin{figure}[hbt]
    \centering
    \includegraphics[width = 0.95\columnwidth, trim={0 0.5cm 0 0.5cm}, clip]{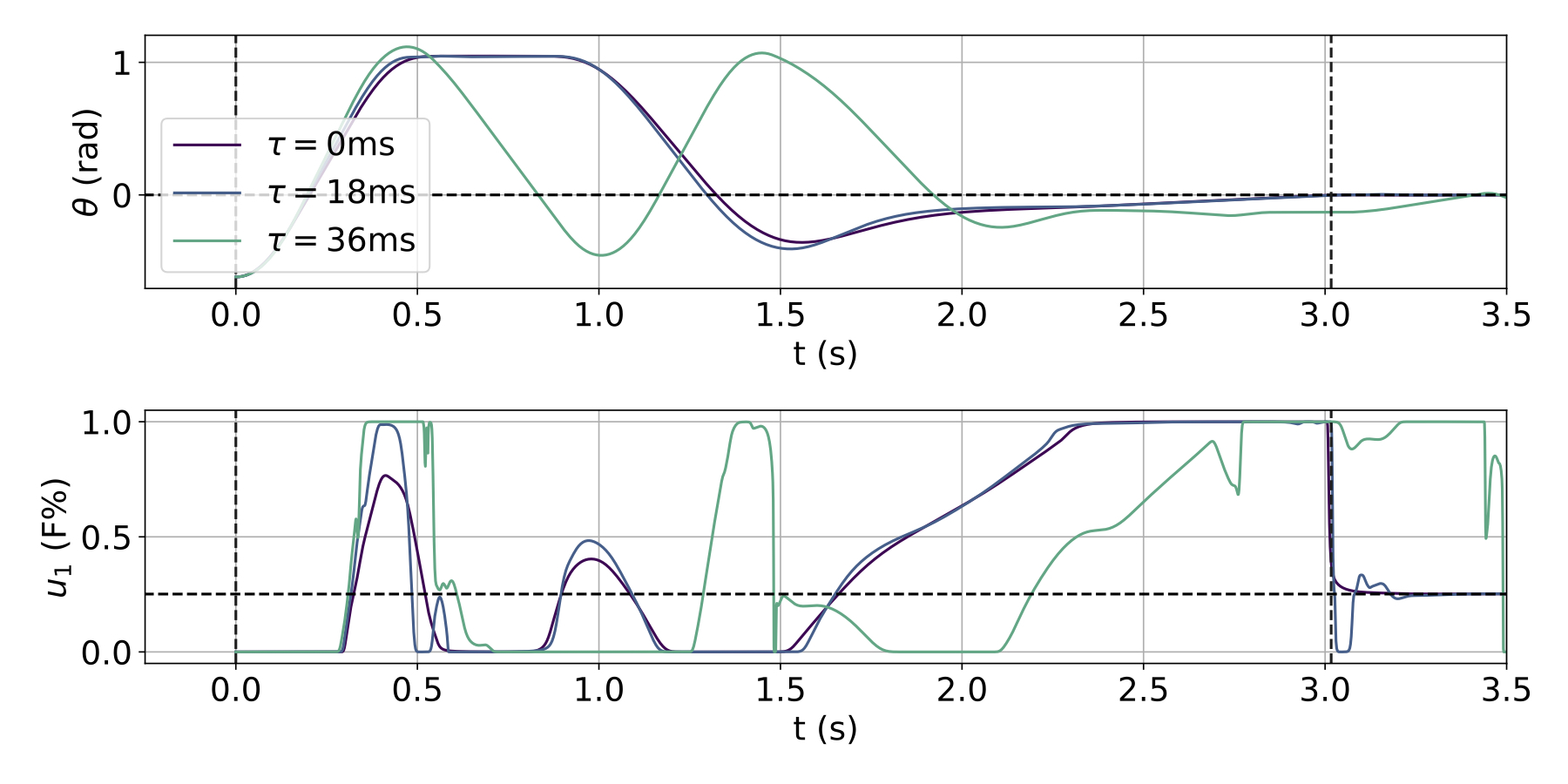}
    \caption{Pitch (top) and the left thrust (bottom) during a G\&CNet driven trajectory simulated with control delays of $0ms$, $18ms$ and $36ms$. The vertical dashed lines show the initial and final time of the true optimal trajectory. The horizontal dashed lines show the target final states: $\theta(t_f)=0.0$ and $u_1(t_f)=u_{\text{hover}}$.}
    \label{fig:stability margin}
\end{figure}

One of the foremost important things is the stability of any controller used on the quadrotor as an unstable controller can lead to failure. The primary stability concerns arise due to the fact that in a real quadrotor there is a measurable delay between the computation of the controls, the state given to the controller and the controller response which arises due to factors such as the time taken to compute the state, and the inertia of the rotors. This delay can be modeled by a fixed time between the command and the execution of the control command:

\begin{equation}
    \mathbf{u}(t) = \mathcal{N}(\mathbf{x}(t-\tau))
\end{equation}

where $\tau$ is the time delay. Using the methods developed in \cite{izzo2018stability}, we find that the stability margin of the G\&CNet($\epsilon=0.2$) controller is $\tau_s=63.8ms$. Although this stability margin is high, it mostly provides information as to the hovering stability of the quadrotor, but we are more interested in the general stability during flight. Fig.~\ref{fig:stability margin} shows the effect of an increasing time delay on the G\&CNet($\epsilon=0.2$) controller left thrust and pitch for delays of $\tau=0ms$, $\tau=18ms$ and $\tau=36ms$. Here we see that, as the delay increases, the controller becomes increasingly unstable up to the point where it is no longer able to track the optimal trajectory nor hover in the final state.

\subsection{Differential flatness based aggressive trajectory generation and control (DiffG\&C)}
A commonly used aggressive trajectory generation method is to use high order polynomials  $P(t) = \mathbf{p}^{\rm T}\mathbf{t}$ to connect the initial point, the waypoints and the final point \cite{mellinger2011minimum,bry2015aggressive}. Thanks to the differential flatness properties of the quadrotor, the thrust on each rotor can be directly related to the $4^{th}$ order derivative of the position curves $\mathbf{u}=\mathbf{f}(\mathbf{p},t)$ \cite{mellinger2011minimum,faessler2017differential}. In particular, in this method, we use the same kinematics model as the reference \cite{faessler2017differential} with Bebop's drag coefficient ($\mathbf D = \mathrm{diag}(-\beta, -\beta, -\beta)$), mass $m=0.389\ \mathrm{kg}$ and length $L=0.08\ \mathrm{m}$.


\begin{align}
\begin{cases}
    \dot{\mathbf{x}} = \mathbf{v} \\
    \dot{\mathbf{v}} = \mathbf{g} + \mathbf{T} + \mathbf{R}^{\rm T}\mathbf{D}\mathbf{R}\mathbf{v} \\
    \dot{\mathbf{\Phi}} = \mathbf{R}'\mathbf{q}
\end{cases}
\label{equ:drone 9 states model}
\end{align}

where thrust $\mathbf{T}= [0, 0, T]^{\rm T}$ and body rate $\mathbf{q}= [p,q,r]^{\rm T}$ are the inputs of the system with the assumption that the low-level acceleration controller and rate controller can track the reference well. Equation~\ref{equ:check input constraints} is used to check the feasibility of the thrust each rotor can provide.

\begin{align}
    \dot{\mathbf{q}} = \mathbf{I}^{-1}(\mathbf{\tau}-\mathbf{q}\times\mathbf{I}\mathbf{q})
    \label{equ:check input constraints}
\end{align}

From the computed polynomial trajectory, the body rate $\mathbf q$ and the rotor thrusts can be determined. For a given arrival time $t_f$, the best trajectory connecting two states is the one with minimal snap. By decreasing the arrival time $t_f$ until the constraints are violated, the polynomial trajectory with minimum arrival time and minimal snap can be found. 

\begin{align}
    \min_{t_f}\Big\{\min_{\mathbf{p}}\int_0^{t_f}P^{(4)}(t)\mathrm{d}t\} &= \min_{t_f}\{\min_{\mathbf{p}}\mathbf{p}^{\rm T}\mathbf{Q}\mathbf{p} \Big\}\label{equ:target}\\
    {\rm s.t.}\;\mathbf{A}\mathbf{p} &= \mathbf{b} \label{equ:constraints} \\
    \mathbf{f}(\mathbf{p},t)&<\mathbf{c} \label{equ:inputs constraints}
\end{align}

where (\ref{equ:target}) is the optimization target, the integral of the $4^{th}$ order derivative of the polynomial which can be written as a quadratic form. Equation~\ref{equ:constraints} is the constraints of the polynomial and (\ref{equ:inputs constraints}) gives the input constraints. The readers are referred to \cite{bry2015aggressive} for the detailed derivation of matrix $\mathbf{Q}$. The algorithm is listed below


\begin{algorithm}
\caption{The pseudocode of DiffG\&C}
\begin{algorithmic}[1]
\Procedure{diff\_control\_guidance}{$t_f,\mathbf{b},\mathbf{c}$}
    \While{$\mathbf{f}(\mathbf{p},t)<\mathbf{c}$}  \Comment{check feasibility}
        \State $\mathbf{p}^* = \mathbf{p}$  
        \State $t_f = t_f - \Delta t$ \Comment{minimise time}
        \State $\min_{\mathbf{p}}\mathbf{p}^{\rm T}\mathbf{Q}\mathbf{p}\quad {\rm s.t.}\;\mathbf{A}\mathbf{p} = \mathbf{b}$ \Comment{gradient descent}
    \EndWhile \\
    \Return $t_f$, $\mathbf{p}^{*}$ \label{roy's loop}
\EndProcedure
\end{algorithmic}
\end{algorithm}


The feed-forward control inputs are computed from the polynomial trajectories and a feedback PID controller is used to compensate for disturbance. The readers are referred to 
\cite{faessler2017differential} for further details on the controller implementation.


We only investigate the movement in the $xoz$ plane by setting any movement in the $y$ direction to $0$. This way, the model given by (\ref{equ:drone 9 states model}) can be simplified to the model in (\ref{equ:simulation model}).

\subsection{Simulation of the G\&CNet Controller}
In this simulation, we use the model from (\ref{equ:simulation model}) as our dynamical model with the rate acceleration $\dot{q}$ and total thrust $T$ as the inputs. The reason is that on the real drone, there are different low-level controllers which can track the thrust and the rate acceleration accurately, one of which is the incremental nonlinear dynamic inversion controller (INDI) \cite{smeur2015adaptive}. We calculate the desired thrust and rate acceleration command from the G\&CNet controller outputs using Eq.~(\ref{equ:calculate cmd from nn output})  

\begin{figure}
    \centering
    \vspace{0.5cm}
    \subfigure[A simulated trajectory using G\&CNet($\epsilon=0.1$).]{\includegraphics[scale=0.07, trim={0 1cm 0 1cm}, clip]{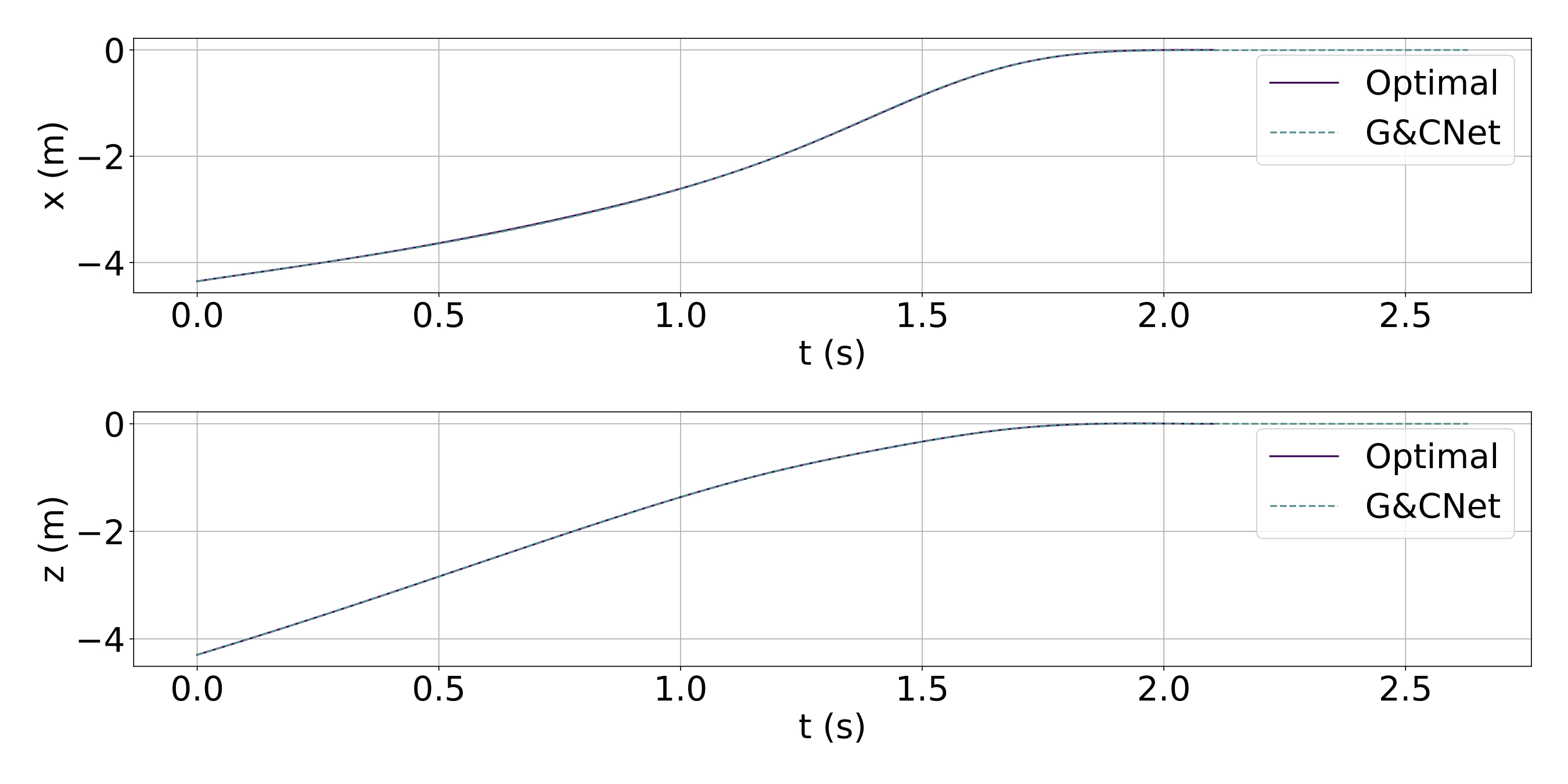}} \\
    \subfigure[Force of the front rotors and the rear rotors along the simulated trajectory.]{\includegraphics[scale=0.07, trim={0 1cm 0 1cm}, clip]{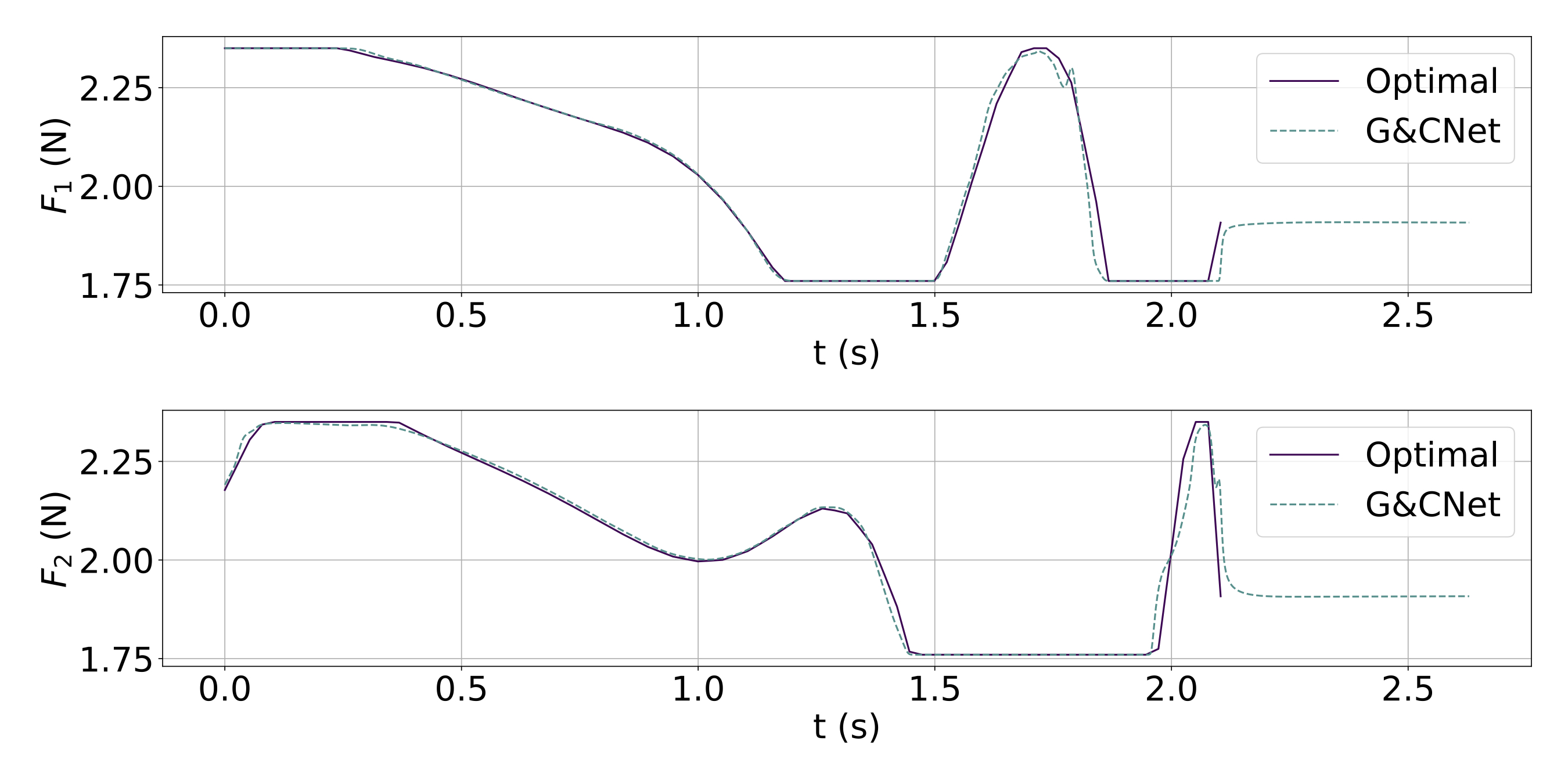}}
    \caption{An example simulation of G\&CNet($\epsilon=0.2$). In each simulation step, the controller receives $x$,$z$,$v_x$,$v_z$,$\theta$,$q$ and outputs the thrust command of the front rotors and the rear rotors. The desired total thrust $T$ and rate acceleration $\dot{q}$ are calculated by (\ref{equ:calculate cmd from nn output}) and sent to model \ref{equ:simulation model} for integration.}
    \label{fig: simulation G&CNet}
\end{figure}

\begin{align}
\begin{cases}
    \dot{q}_{cmd}=\frac{(u_1-u_2)\Delta FL}{I_{xx}} \\ T_{cmd} = \frac{(u_1+u_2)\Delta F}{m}
\end{cases}
\label{equ:calculate cmd from nn output}
\end{align}


\subsection{Comparison between DiffG\&C and G\&CNet}
\begin{figure}
    \centering
    \vspace{0.5cm}
     \subfigure[$\epsilon=0.5$]{\includegraphics[width = 0.45\columnwidth, trim={0.2cm 0.9cm 0.8cm 0.9cm}, clip]{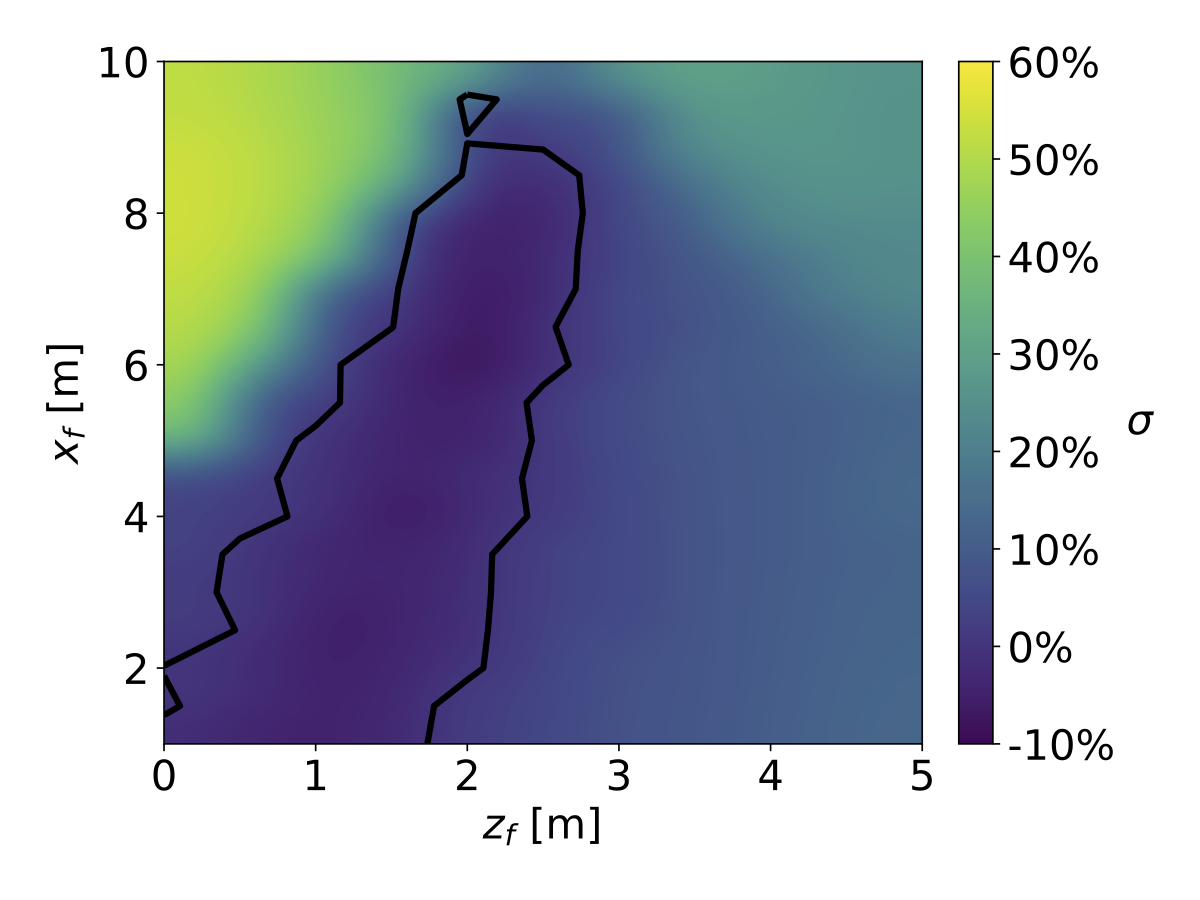}} 
     \subfigure[$\epsilon=0.2$]{\includegraphics[width = 0.45\columnwidth, trim={0.2cm 0.9cm 0.8cm 0.9cm}, clip]{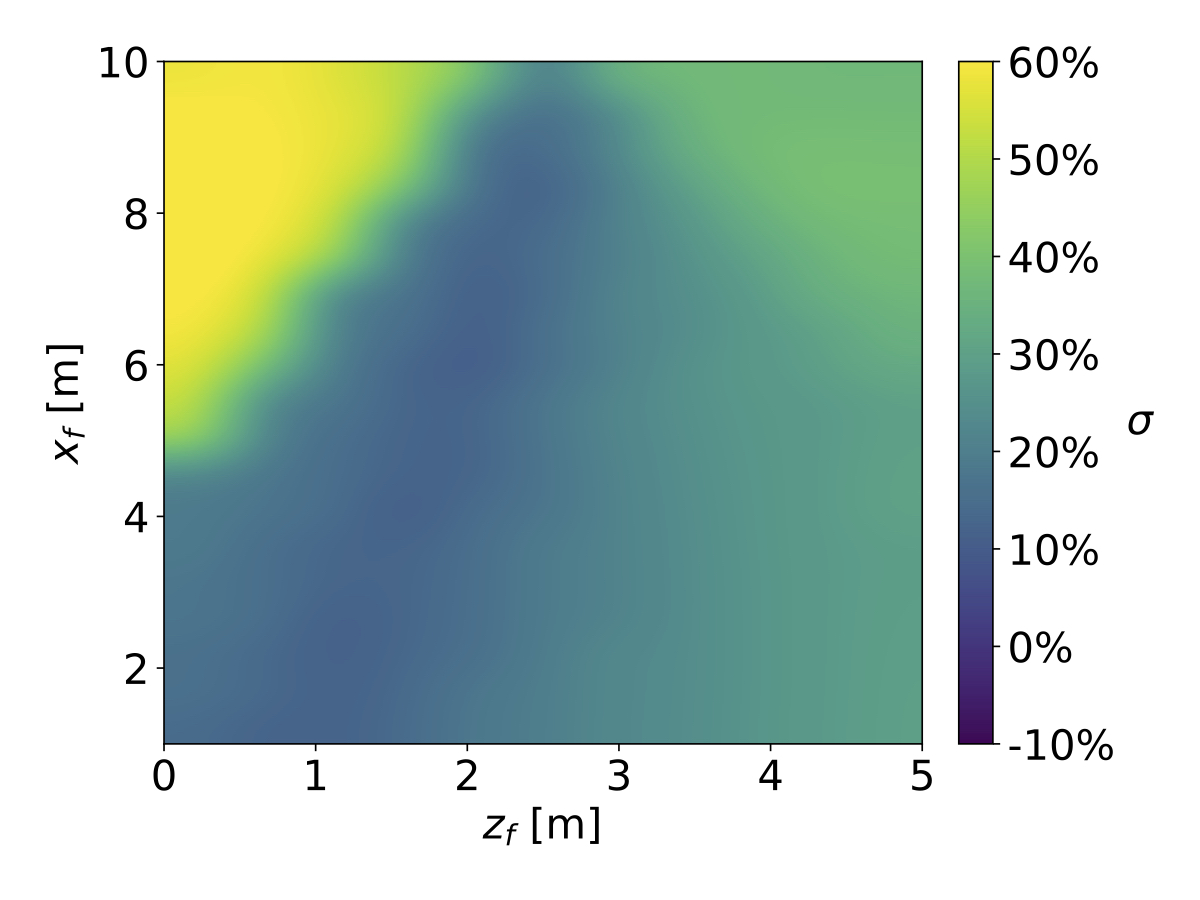}} 
    \caption{Comparison of arrival time between DiffG\&C and G\&CNet. Despite power optimality being weighted equally to time optimality, G\&CNet($\epsilon=0.5$) can, in most cases, steer the drone to the target points in less time than DiffG\&C (the black line shows the region border where G\&CNet outperforms DiffG\&C). On the other hand, G\&CNet($\epsilon=0.2$) is always faster than DiffG\&C.}
    \label{fig: OPT vs OOC} 
\end{figure}

In this section, a comparison is made in simulation between DiffG\&C and G\&CNet. The time required by the drone to reach the target is used to derive a performance index. In each trial, the initial position of the drone is set to be $[x_0,z_0] = [0m,2.5m]$ and the same target $x_f\in[1,10]$, $z_f\in[0,5]$ is set for both controllers. To quantify the performance of a method, we introduce an index $\sigma$:

\begin{align}
    \sigma = \frac{t_f^{DiffG\&C}-t_f^{G\&CNet}}{t_f^{DiffG\&C}}
\end{align}

where $t_f^*$ is the arrival time of each controller. When $\sigma>0$, the G\&CNet controller is faster than DiffG\&C and vice versa. Fig.~\ref{fig: OPT vs OOC} gives the simulation results of multiple target points with $\epsilon=0.2$ and $\epsilon=0.5$. From Fig.~\ref{fig: OPT vs OOC}(a), it can be seen that, in most cases, G\&CNet($\epsilon=0.5$) has a shorter arrival time than DiffG\&C outside the region delineated by the black border, and in this region the arrival time is within 10\% of DiffG\&C. As seen in Fig.~\ref{fig: OPT vs OOC}(b), with G\&CNet($\epsilon=0.2$), the arrival time is always shorter and up to 60\% faster than DiffG\&C.


\begin{figure}
  \centering
    \subfigure[]{\includegraphics[width=0.85\columnwidth, trim={0 1cm 0 1cm}, clip]{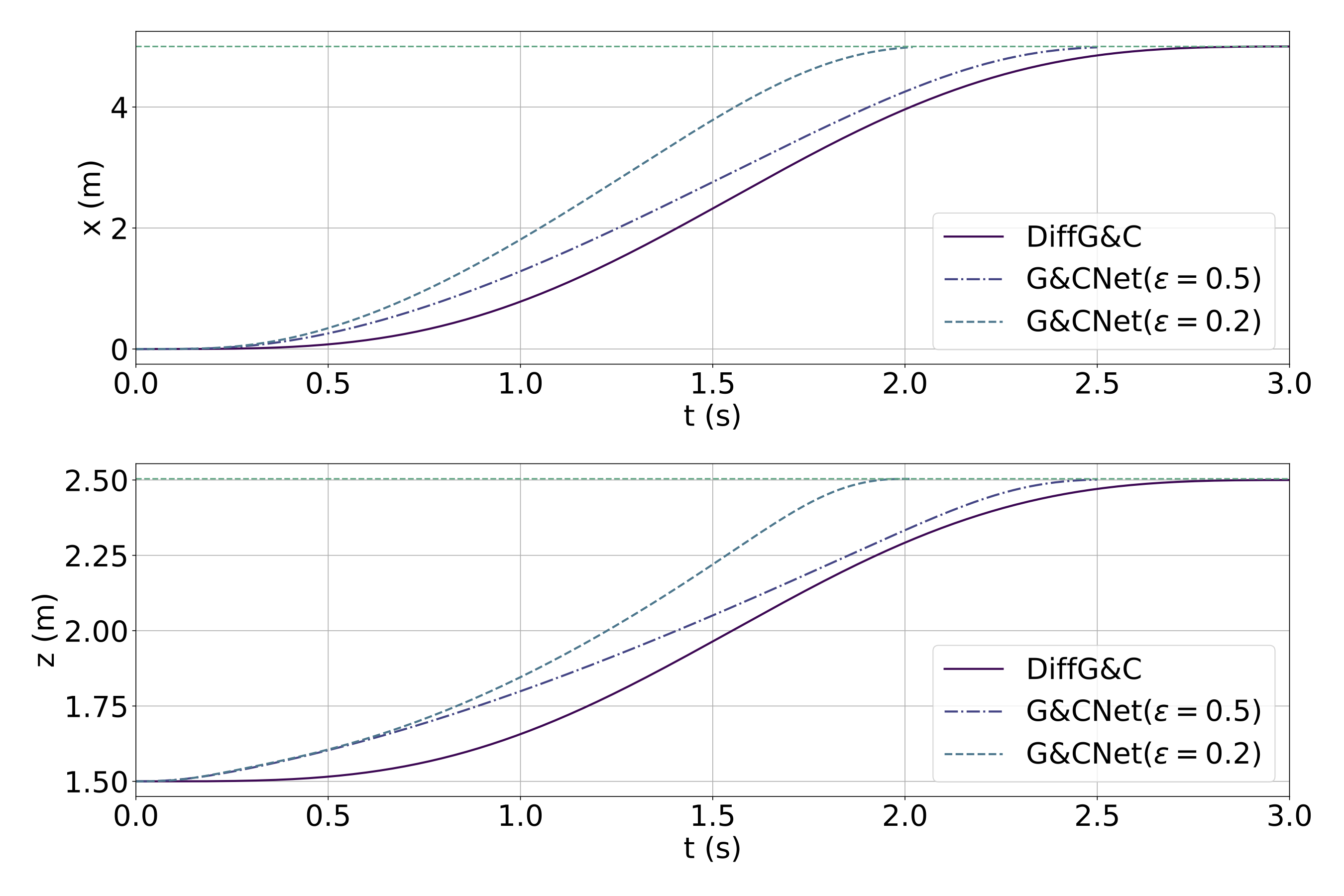}} 
    \subfigure[]{\includegraphics[width=0.85\columnwidth, trim={0 1cm 0 1cm}, clip]{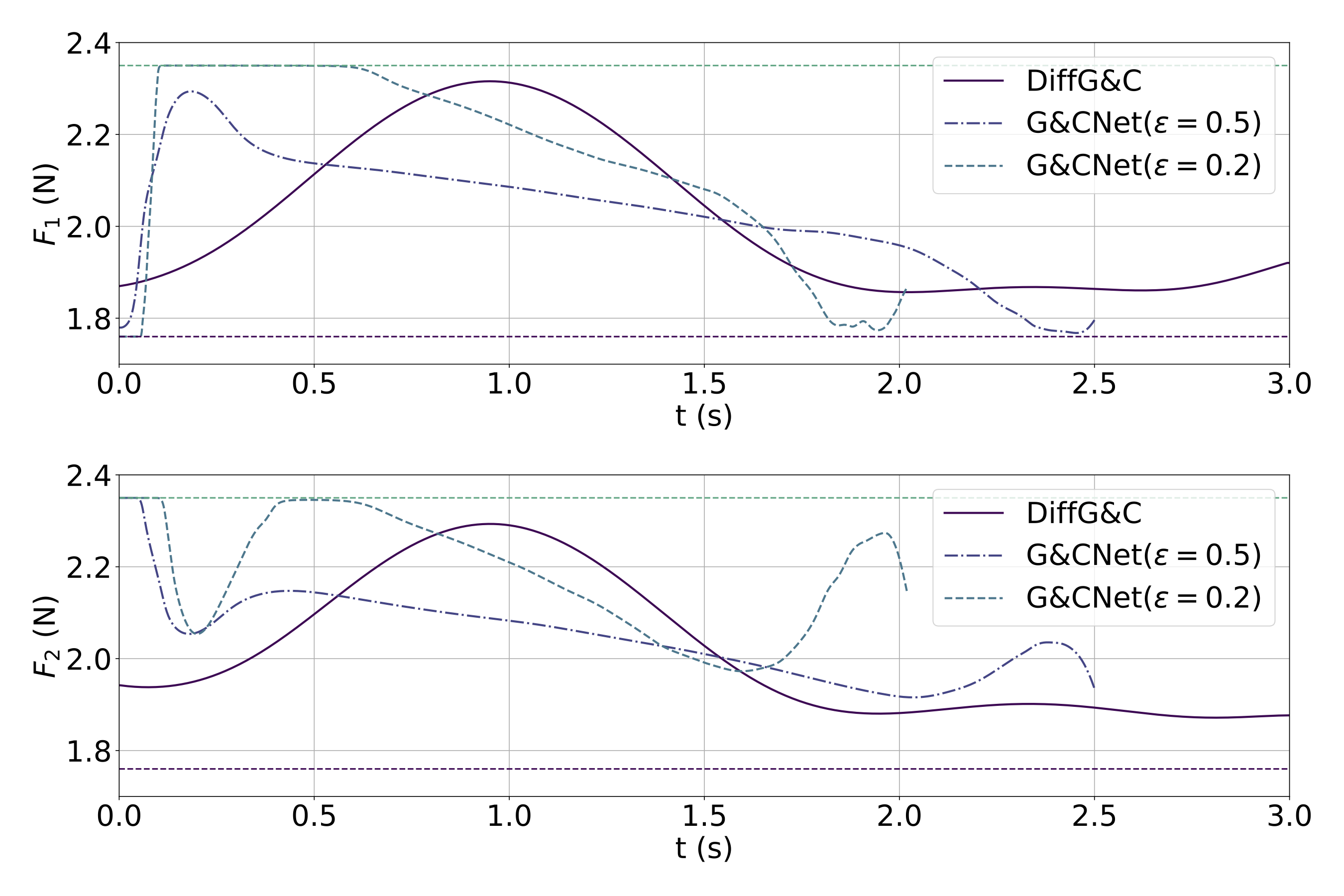}}
    \caption{An example of comparison between DiffG\&C and G\&CNet($\epsilon=0.5$) when $x_f=5$, $z_f=2.5$.}
    \label{fig:trajectory example}
\end{figure}

Fig.~\ref{fig:trajectory example} shows a comparison plot of the trajectories and controls of DiffG\&C, G\&CNet($\epsilon=0.5$) and G\&CNet($\epsilon=0.2$). It can be seen that all three controllers reach the target, but the control profiles and arrival times differ significantly. With DiffG\&C, due to the polynomial representation of trajectories, the quadrotor inputs cannot be fully utilised, and thus the time-optimality cannot be guaranteed. On the other hand, G\&CNets are able to saturate the inputs and arrive at a similar or smaller time.


\section{EXPERIMENT SETUP AND RESULT}
In this section we show the experimental setup for real-world flights and the flight performance of each method. 

\subsection{Experiment Setup}

\begin{figure}
    \centering
    \includegraphics[width=0.6\columnwidth, trim={0 12cm 0 21cm}, clip]{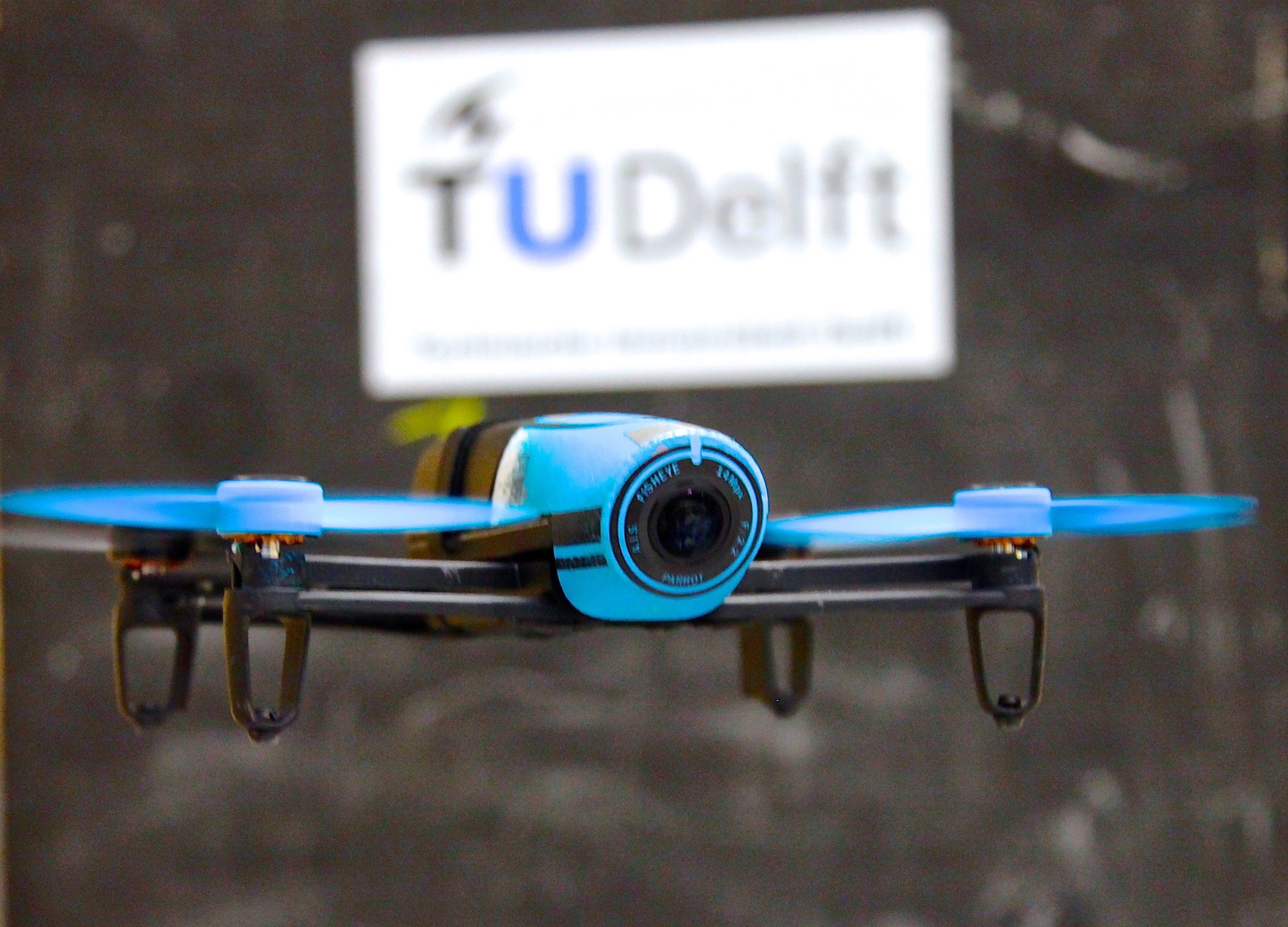} \\
    \caption{A Parrot Bebop 1 is used as the flying platform. The original autopilot is fully replaced by an open-source autopilot called Paparazzi UAV.}
    \label{fig: bebop} 
\end{figure}

\begin{figure}
    \centering
      \subfigure[The control structure of G\&CNet. A PD controller and 
      \newline an INDI controller are used to keep the quadrotor at \newline $y=0m$ and $\psi=0^{\circ}$.]{ \includegraphics[width=0.85\columnwidth]{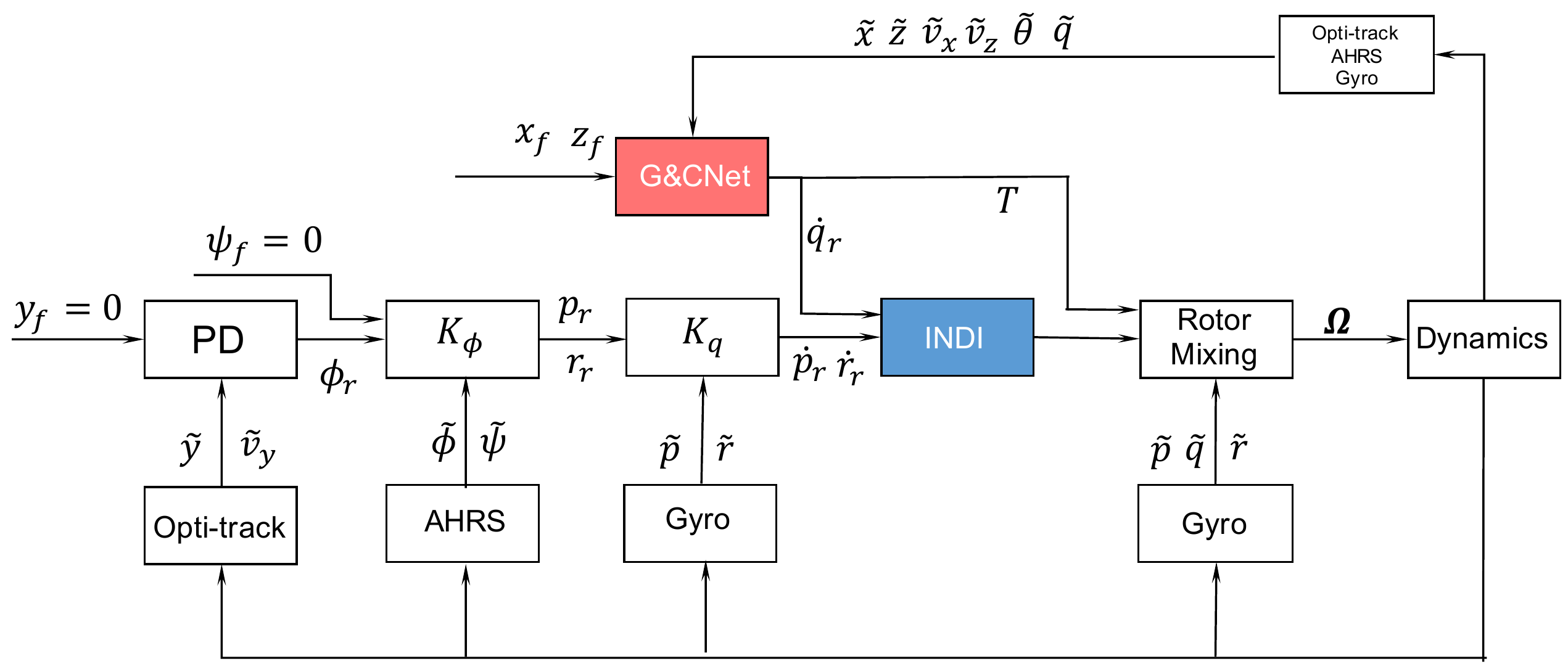} } 
    \centering
      \subfigure[The control structure of DiffG\&C. The feed-forward signal is \newline directly computed from generated trajectories. A feedback \newline controller is used to correct for deviations.]{ \includegraphics[width=0.85\columnwidth]{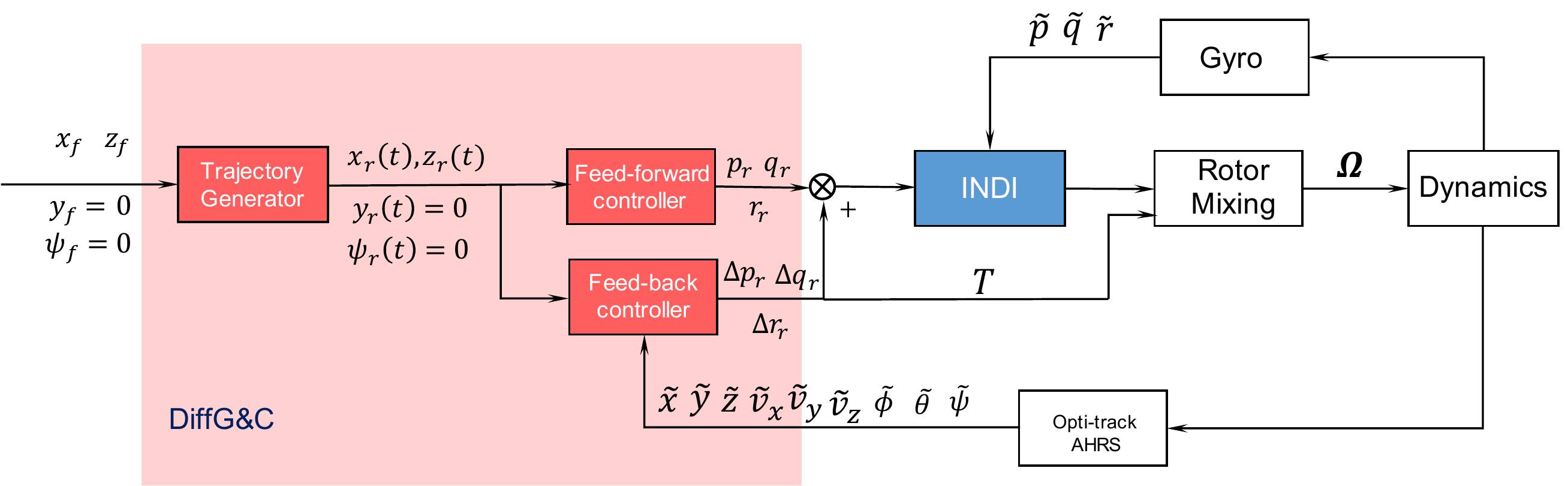} }
    \caption{The control structure of the proposed G\&CNet and the benchmark DiffG\&C.}
    \label{fig: control structure} 
\end{figure}

To verify the proposed G\&CNet, we use a commercial Parrot Bebop 1 as our flying platform (Fig.~\ref{fig: bebop}). The This fully replaced by an open-source autopilot, Paparazzi-UAV. This autopilot provides full access to the raw sensor data and rotor commands.
In this experiment, the position and velocity feedback are from Opti-track motion capture system. The attitude estimation is from an on-board complementary filter, which is inevitably biased. The angular rate estimation is from the on-board gyroscope. The control architecture is shown in Fig.~\ref{fig: control structure}. For G\&CNet, the lateral movement and heading are controlled by the original outer-loop PID controller and inner-loop INDI controller to keep $y=0$ and $\psi=0^{\circ}$. The maneuver on the vertical plane is taken over by the proposed G\&CNet. In each control update, G\&CNet receives the state estimations and outputs the desired pitch acceleration $\dot{q}$ and the thrust $T$. For the benchmark DiffG\&C, after the trajectory is generated, the desired angular rate $\mathbf{q}$ can be directly calculated. Then a feedback controller is used to compensate the deviation caused by the model inaccuracy, and the state estimation bias.

In the real-world flight tests, we test $3$ controllers which are DiffG\&C, G\&CNet($\epsilon=0.5$) and G\&CNet($\epsilon=0.2$) respectively. For each controller, the start position is set to be $\mathbf{x}_0=[0m,-1.5m]^{\rm T}$ and $3$ targets which are $\mathbf{x^1_f}=[5m,2.5m]^{\rm T}$,$\mathbf{x^2_f}=[5m,1.5m]^{\rm T}$ and $\mathbf{x^3_f}=[5m,0.5m]^{\rm T}$ are set to be tested. For each target, we have $10$ independent flights. To evaluate the performance of one controller, we have $2$ indices which are average arrival time $\Delta \Bar{t}_*$ and average tracking error $\Delta\Bar{x}_*$ defined by

\begin{align}
    \Delta \Bar{t}_* &=\frac{\sum_i^N\Delta t^i_*}{N} \\
    \Delta \Bar{x}_* &= \frac{\sum_{i=1}^{N}\sum_{j=1}^{n_i}\norm{\hat{\mathbf{x}}^{i,j}_*-{\mathbf{x}_{r}^{i,j}}_*}}{\sum_{i=1}^Nn_i}
\end{align}

\begin{figure*}[hbt]
    \centering
    \vspace{0.2cm}
     \subfigure[
     DiffG\&C]{\includegraphics[scale = 0.4,trim={4cm 8cm 4cm 9cm},clip]{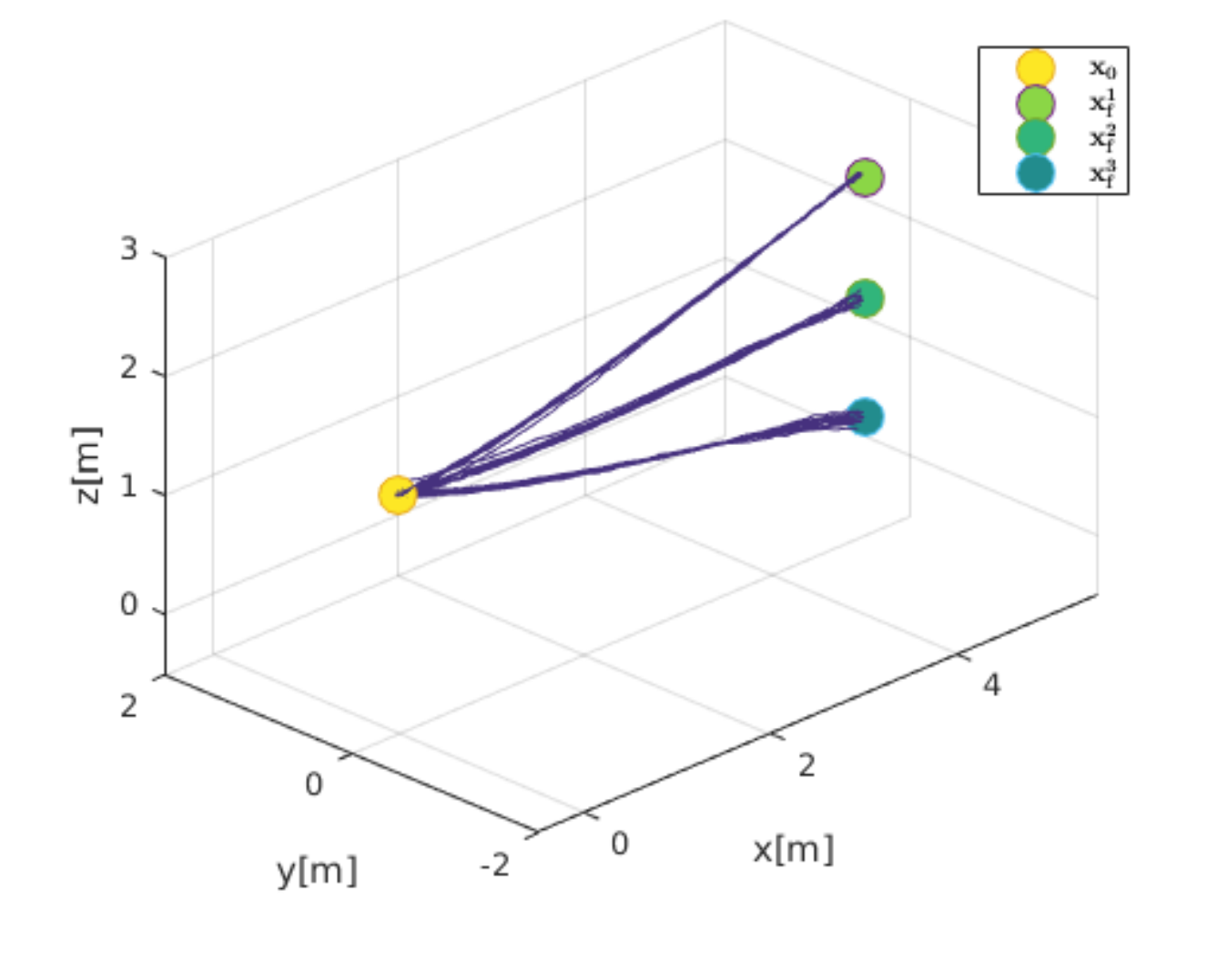}} 
     \subfigure[
     GCNet($\epsilon=0.5$)]{\includegraphics[scale = 0.4,trim={4cm 8cm 4cm 9cm},clip]{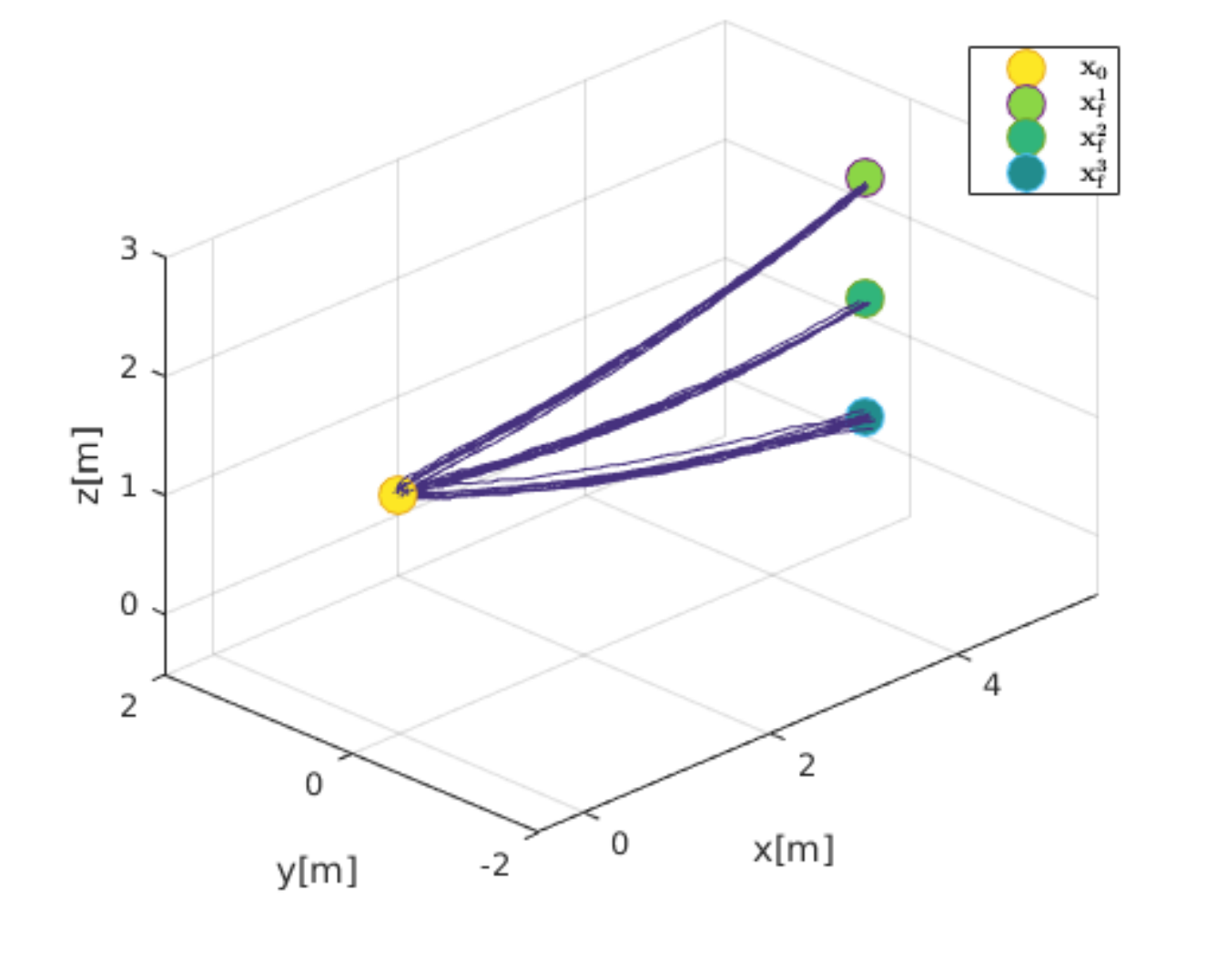}} 
     \subfigure[
     GCNet($\epsilon=0.2$)]{\includegraphics[scale = 0.4,trim={4cm 8cm 4cm 9cm},clip]{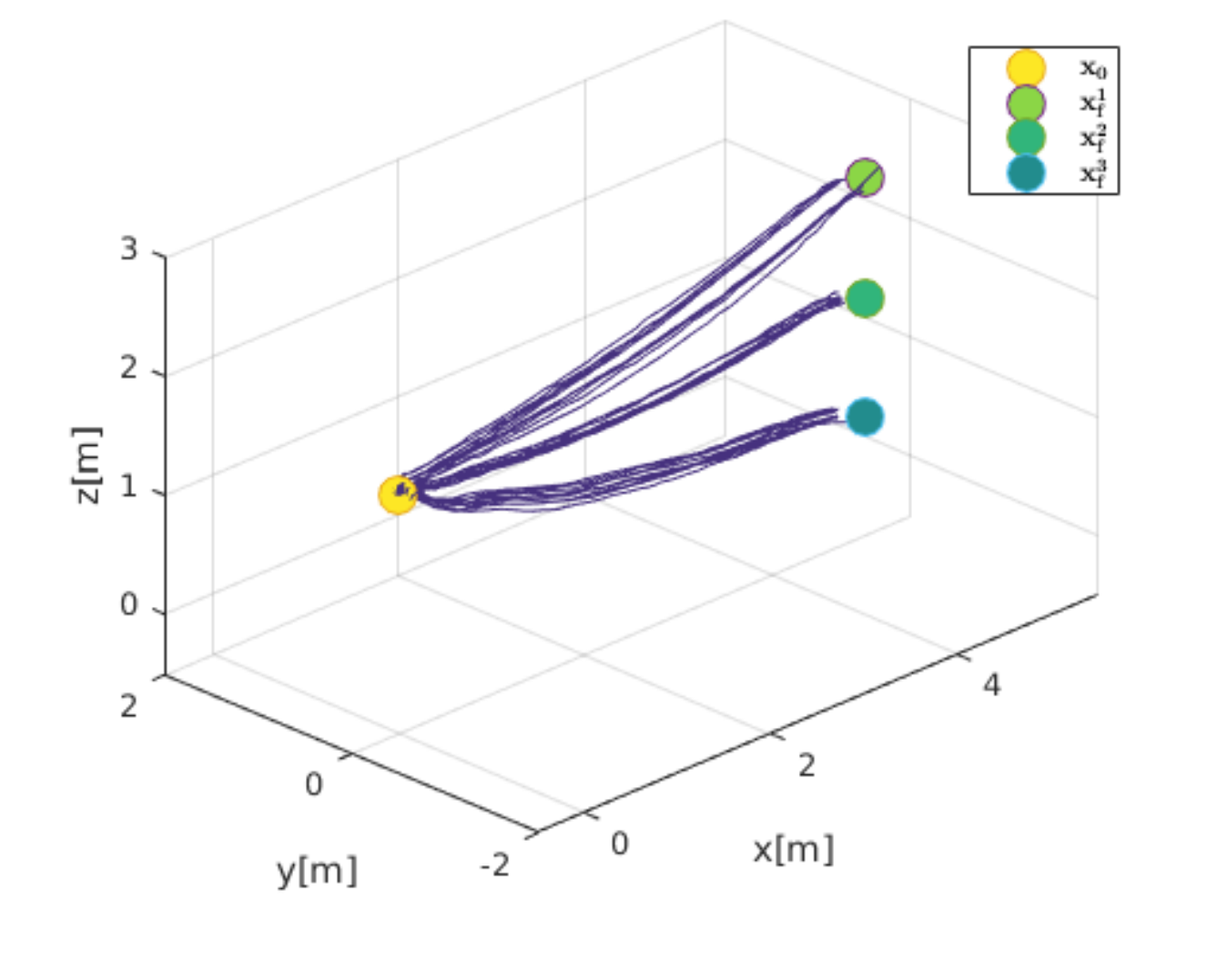}} 
    \caption{The real-world flight data of different controllers to different targets}
    \label{fig: real-world flight data} 
\end{figure*}

where $\Delta t_*^i$ is the arrival time of $i^{th}$ flight of method $*$. $N$ is the number of the flight of one controller, which is $10$ in our case. $\hat{\mathbf{x}}^{i,j}_*$ is the position of $i^{th}$ flight's $j^{th}$ sample measured by the Opti-track system. ${\mathbf{x}_{r}^{i,j}}_*$ is the corresponding position reference. It should be noted that in DiffG\&C, $\mathbf{x}_r$ is the reference trajectory while in G\&CNet, it is the simulated trajectory.

\subsection{Experiment Result}
The experiment is set up as described in the previous section and we have $90$ flights in total ($3$ controllers $\times$ $3$ targets $\times$ $10$ flights, depicted in Fig.~\ref{fig: real-world flight data}). The average arrival time is listed in Table~\ref{tab:average arrival time} and the average tracking error is listed in Table~\ref{tab:average tracking error}.

\begin{table}[h]
\caption{Average arrival time $\Delta\Bar{t}_*$ to targets $\mathbf{x^i_f}$ 
}
\label{tab:average arrival time}
\begin{center}
\begin{tabular}{|c||c|c|c|}
\hline
  \textbf{Controller}  & $\mathbf{x^1_f}$ &  $\mathbf{x^2_f}$ & $\mathbf{x^3_f}$ \\
\hline
DiffG\&C & $2.63s\pm0.05s$  &  $2.18s\pm0.02s$  & $2.10s\pm0.04s$ \\
\hline
G\&CNet($
0.5$) & $2.36s\pm0.02s$  &  $2.20s\pm0.02s$  & $2.13s\pm0.01s$ \\
\hline
G\&CNet($
0.2$) & $1.96s\pm0.03s$  &  $1.88s\pm0.03s$  & $1.91s\pm0.04s$ \\
\hline
\end{tabular}
\end{center}
\end{table}

\begin{table}[h]
\caption{Average tracking error $\Delta \Bar{x}_*$ to targets $\mathbf{x^i_f}$}
\label{tab:average tracking error}
\begin{center}
\begin{tabular}{|c||c|c|c|}
\hline
  \textbf{Controller}  & $\mathbf{x^1_f}$ &  $\mathbf{x^2_f}$ & $\mathbf{x^3_f}$ \\
\hline
DiffG\&C                     & $0.06m$  &  $0.07m$  & $0.07m$ \\
\hline
G\&CNet($\epsilon=0.5$) & $0.13m$  &  $0.09m$  & $0.10m$ \\
\hline
G\&CNet($\epsilon=0.2$) & $0.17m$  &  $0.15m$ & $0.28m$ \\
\hline
\end{tabular}
\end{center}
\end{table}

From Table~\ref{tab:average arrival time}, it can be seen that when the target is set to $\mathbf{x^1_f}$, G\&CNet($\epsilon=0.5$) reaches the target in a shorter time DiffG\&C, whereas for targets $\mathbf{x^2_f}$ and $\mathbf{x^3_f}$, it is on par with the benchmark. On the other hand, G\&CNet($\epsilon=0.2$) always reaches the target in faster time. These experimental results confirm the simulation results that were obtained in Section~\ref{sec:SIMULATION RESULT AND ANALYSIS}.


In terms of tracking error, DiffG\&C has the smallest tracking error $\Delta \Bar{x}$ 
followed by G\&CNet($\epsilon=0.5$), and finally G\&CNet($\epsilon=0.2$). We find that G\&CNet($\epsilon=0.5$) outperforms G\&CNet($\epsilon=0.2$) in terms of the tracking error. This can be attributed to the fact that a lower $\epsilon$ corresponds to a more aggressive trajectory and, in turn, a high-frequency high amplitude changes of the inputs. As mentioned in Section~\ref{sec:optimalcontrol}, this is difficult for the quadrotor to track due to the inertial properties of its rotors.

\section{CONCLUSIONS}
We have proposed G\&CNet as a novel online optimal controller for quadrotors that removes the need for expensive real-time optimal trajectory generation by learning a deep neural representation of the optimal state-control mapping. We have demonstrated, both in simulation and with real-world flight tests, that G\&CNets are not only feasible for this purpose, but also competitive with a commonly used method, DiffG\&C. Our results indicate that a G\&CNet weighting equally power and time optimality ($\epsilon=0.5$) is, at worst, 10\% slower than DiffG\&C and faster most of times while a G\&CNet aggressively biased towards time optimality ($\epsilon=0.2$) is always considerably faster by up to 60\%.

There are many avenues of exploration available. Future work can focus on adding the actuator model into the optimal control problem thus eliminating the issue of difficult to track bang-bang controls for the rotors. A further extension of our work would be to implement the optimal control problem in the full 3-dimensional model thus potentially adding more interesting manoeuvre capabilities to the quadrotor. Additionally, the network could be trained to achieve a nonzero velocity in the final state in preparation for consecutive manoeuvres.



\addtolength{\textheight}{-16cm}   





\bibliography{IEEEexample}
\bibliographystyle{ieeetr}



\end{document}

%% file: Figures/axisdef2.pdf_tex

\begingroup
  \makeatletter
  \providecommand\color[2][]{%
    \errmessage{(Inkscape) Color is used for the text in Inkscape, but the package 'color.sty' is not loaded}
    \renewcommand\color[2][]{}%
  }
  \providecommand\transparent[1]{%
    \errmessage{(Inkscape) Transparency is used (non-zero) for the text in Inkscape, but the package 'transparent.sty' is not loaded}
    \renewcommand\transparent[1]{}%
  }
  \providecommand\rotatebox[2]{#2}
  \ifx\svgwidth\undefined
    \setlength{\unitlength}{425.2pt}
  \else
    \setlength{\unitlength}{\svgwidth}
  \fi
  \global\let\svgwidth\undefined
  \makeatother
  \begin{picture}(1,0.37629351)%
    \put(0,0){\includegraphics[width=\unitlength]{Figures/axisdef2.pdf}}%
    \put(0.14446939,0.34014687){\color[rgb]{0,0,0}\makebox(0,0)[lb]{\smash{$u_1$}}}%
    \put(0.90343625,0.25284009){\color[rgb]{0,0,0}\makebox(0,0)[lb]{\smash{$u_2$}}}%
    \put(0.09675952,0.14953072){\color[rgb]{0,0,0}\makebox(0,0)[lb]{\smash{$\theta$}}}%
    \put(0.25455193,0.05038961){\color[rgb]{0,0,0}\makebox(0,0)[lb]{\smash{$X$}}}%
    \put(0.00007444,0.23088306){\color[rgb]{0,0,0}\makebox(0,0)[lb]{\smash{$Z$}}}%
    \put(0.57824977,0.26117621){\color[rgb]{0,0,0}\makebox(0,0)[lb]{\smash{$v_z$}}}%
    \put(0.66012319,0.14528799){\color[rgb]{0,0,0}\makebox(0,0)[lb]{\smash{$v_x$}}}%
    \put(0.42618848,0.00543349){\color[rgb]{0,0,0}\makebox(0,0)[lb]{\smash{$q$}}}%
  \end{picture}%
\endgroup